\documentclass{article}

     \PassOptionsToPackage{numbers, compress}{natbib}
   \usepackage[preprint]{neurips_2019}

\usepackage[utf8]{inputenc} % allow utf-8 input
\usepackage[T1]{fontenc}    % use 8-bit T1 fonts
\usepackage{hyperref}       % hyperlinks
\usepackage{url}            % simple URL typesetting
\usepackage{booktabs}       % professional-quality tables
\usepackage{amsfonts}       % blackboard math symbols
\usepackage{nicefrac}       % compact symbols for 1/2, etc.
\usepackage{microtype}      % microtypography
\usepackage{graphicx}
\usepackage{amsmath}

\title{Learn to Estimate Labels Uncertainty for Quality Assurance}

\author{%
  Agnieszka Tomczack \\
  Technical University of Munich\\
  \texttt{agnieszka.tomczak@tum.de} \\
   \And
   Nassir Navab \\
 Technical University of Munich\\
  \texttt{nassir.navab@tum.de} \\
  \And
   Shadi Albarqouni \\
 Technical University of Munich\\
  \texttt{shadi.albarqouni@tum.de} \\  \\
}

\begin{document}

\maketitle

\begin{abstract}
Deep Learning sets the state-of-the-art in many challenging tasks showing outstanding performance in a broad range of applications. Despite its success, it still lacks robustness hindering its adoption in medical applications. Modeling uncertainty, through Bayesian Inference and Monte-Carlo dropout, has been successfully introduced for better understanding the underlying deep learning models. Yet, another important source of uncertainty, coming from the inter-observer variability, has not been thoroughly addressed in the literature.  In this paper, we introduce \textit{labels uncertainty} which better suits medical applications and show that modeling such uncertainty together with epistemic uncertainty is of high interest for quality control and referral systems.
\end{abstract}

\section{Introduction}
\label{sec:intro}

Deep Learning (DL) has emerged as a leading technology for accomplishing many challenging tasks showing outstanding performance in a broad range of applications~\cite{lecun2015deep}. 
However, despite its success and merit in recent state-of-the-art methods, DL has so far been suggested without any risk-management hindering its adoption in medical applications~\cite{cabitza2017unintended,miotto2017deep}.
For instance, clinicians and medical experts cannot accept predictions as blind faith; however, they would like to understand the reasoning behind the decision made in terms of interpretability, and trustworthiness, \emph{i.e.} confidence level and source of uncertainties. This demands explainable AI models that report the associated uncertainty alongside the prediction enabling the physicians to screen such samples for final diagnosis.

\begin{figure}
	\label{fig:teaser}
	\includegraphics[width=\textwidth]{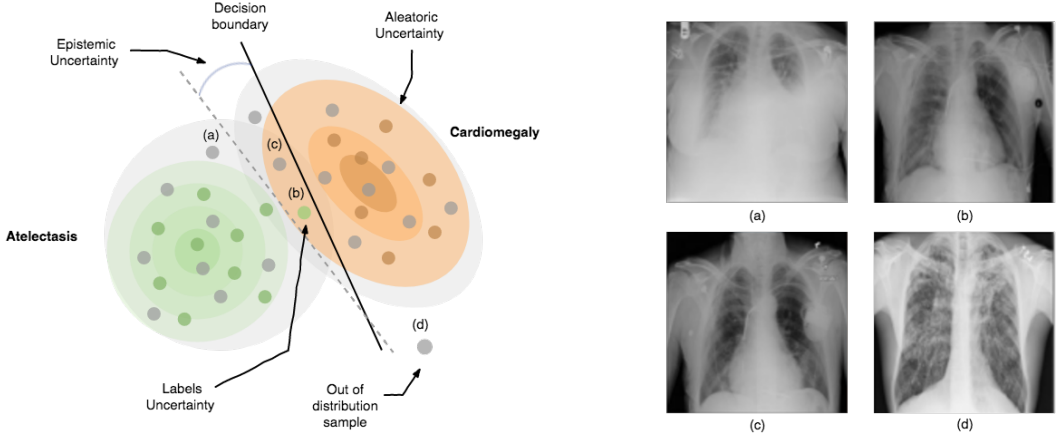}
	\caption{An illustrative example of different sources of uncertainties in Medical Image Analysis. Left: Embedding of the latent features of CXR images and their corresponding labels. Right: CXR images with different diagnosis; Atelectasis (a), Cardiomegaly (b,c), and Fibrosis (d).}
\end{figure}

Modeling such uncertainty in Neural Networks was investigated back in the '90s by Nix and Weigend~\cite{nix1994estimating} where the mean and variance of the probabilistic distribution were estimated and used as a measure of uncertainty of the predicted output. A few years later, Tom Heskes~\cite{heskes1997practical} proposed a method to compute the prediction intervals, using the variance and the width of the confidence interval, which is necessary for models trained with a limited amount of data. Recently, Gal \emph{et al.}~\cite{gal2016dropout} and Kendall \emph{et al.}~\cite{kendall2017uncertainties} propose Monte-Carlo dropout (MCDO) in the context of Bayesian Inference to model both \textit{epistemtic} (model uncertainty) and \textit{aleatoric} (variance in the data distribution) uncertainties in DL (\emph{cf.} Fig.\ref{fig:teaser}). Their MCDO approximation has been successfully applied in many medical applications~\cite{leibig2017leveraging,jungo2018effect,chen2018deep} providing the clinicians with a valuable tool for quality control.

Another source of uncertainty, so-called \textit{labels} uncertainty, coming from the "noisy" labels due to the high variability among the raters, is argued to be of high importance in clinical setup~\cite{guan2018said}. 
One approach to the problem would be modeling inter-observer variability through crowd-sourcing techniques, as suggested in ~\cite{albarqouni2016aggnet, guan2018said}, and interpret the variance of multiple predictions as labels uncertainty. However, the relation to the epistemic uncertainty of the model remains unclear.  

Jungo \emph{et al.}~\cite{jungo2018effect} has recently investigated modeling the inter-observer variability showing that the learned \textit{labels} uncertainty can be implicitly combined with the \textit{epistemic} uncertainty. In other words, training the model with noisy labels would be reflected in the model's parameters. Having said that, Leibig \emph{et al.}~\cite{leibig2017leveraging} raised a valid point in their discussion regarding the ambiguity of the uncertainty source, specifically, the \textit{epistimic} and \textit{labels} uncertainty. A very recent relevant work by Kohl \emph{et al.}~\cite{kohl2018probabilistic} has modeled, however, the \textit{labels} uncertainty in his segmentation framework suggesting modeling such uncertainty is better than modeling \textit{epistimic} uncertainty.

In this paper, we introduce a novel framework to model both \textit{epistimic} and \textit{labels} uncertainty showing one can not replace the other, and both capture different sources of uncertainty and can be fairly distinguished. 
Such a framework would have a significant impact on the decision referral systems by augmenting the physicians with the amount and source of uncertainty for a given query.

\section{Methodology}
\label{sec:method}
Given a dataset of input images $\vec{X}=\{\vec{x}_1, \dots,\vec{x}_N\} \in \mathbb{R}^{n \times N}$, and their corresponding labels annotated by $k$ clinicians $\vec{Y}=\{\vec{y}_1^{(1)}, \dots, \vec{y}_1^{(k)},\dots,\vec{y}_N^{(k)}\} \in \mathbb{R}^{c \times kN}$, our aim is to build a model Neural Network model, 
$\hat{\vec{y}_i} = \vec{f}^{\vec{w}}(\vec{x}_i) + \eta$, 
where $\vec{w}$ is the model parameters, $\eta \sim \mathcal{N}(0, \sigma_{\eta}^2)$ is the measurement noise, and $\sigma_{\eta}^2$ is the noise variance, which captures the \textit{labels} uncertainty, \emph{i.e.} the inter-observer variability. 
To realize that, we follow the theoretical foundations of Bayesian Neural Networks to find the posterior distribution $p(\vec{z_y}|\vec{X},\vec{Y})$ over the latent variable of interest $\vec{z_y}$, \emph{i.e.} the inter-observer variability, given the prior distribution $p(\vec{z_y}|\vec{X})$ as 
\begin{equation}
p(\vec{y}_i| \vec{x}_i,\vec{X},\vec{Y}) = \int\displaylimits_{z_y} p(\vec{y}_i|\vec{x}_i, \vec{z_y})p(\vec{z_y}|\vec{X},\vec{Y})\,d\vec{z_y},
\end{equation}
where $p(\vec{y}_i|\vec{x}_i, \vec{z_y})$ is the probability of a given image $\vec{x}_i$ and a latent code $\vec{z_y}$ representing one annotator. 
The former likelihood would give better understanding of the labels uncertainty at a single weight estimate~\cite{kohl2018probabilistic}. However, to distinguish between \textit{epistemic} and \textit{labels} uncertainties, we need to model the weights uncertainty as well. Under certain conditions, the likelihood can be written as
\begin{equation}
p(\vec{y}_i| \vec{x}_i,\vec{X},\vec{Y}) = \int\displaylimits_{z_y}\int\displaylimits_{w} p(\vec{y}_i|\vec{x}_i, \vec{w}, \vec{z_y})p(\vec{z_y}|\vec{X},\vec{Y}) p(\vec{w}|\vec{X},\vec{Y})\,d\vec{w}\,d\vec{z_y}. 
\end{equation}
Under the assumption of independent variables, the ELBO can be derived similar to App.F in~\cite{kingma2013auto} as 
\begin{equation}
\label{eq:LU}
\begin{split}
%ELBO(q,\theta)    & =  \int q_{\theta}(\vec{w}) p(\vec{z_y}|\vec{X}) \log p(\vec{Y}|\vec{X}, \vec{w}, \vec{z_y}) d\vec{w}d\vec{z_y} - \text{KL}[q_{\theta}(\vec{w})|| p(\vec{w})]  - \beta~\text{KL}[p(\vec{z_y}|\vec{X})|| p(\vec{z_y}|\vec{X},\vec{Y})]\\
ELBO(q,\theta, \phi) & = \mathbb{E}_{\vec{w}, \vec{z_y}}[\log p(\vec{Y}|\vec{X}, \vec{w},\vec{z_y})] - \text{KL}[q_{\theta}(\vec{w})|| p(\vec{w})] \\ 
& - \beta~\text{KL}[p_{\phi}(\vec{z_y}|\vec{X})|| p(\vec{z_y}|\vec{X},\vec{Y})],
\end{split}
\end{equation}
where $\theta$, and $\phi$ are the network parameters, and $\beta$ is a parameter needed in the realm of neural networks~\cite{graves2011practical,blundell2015weight}. The objective function can be then formulated as 
\begin{equation}
\label{LabelsUncertainty}
\begin{split}
\mathcal{L}_{\text{LU}}(\theta, \phi, d) = & -\frac{1}{N} \sum_{i=1}^{N} \log p(\vec{y}_i|\vec{f}^{\hat{\vec{w}_i}}(\vec{x}_i, \vec{z_y}_i)) + \frac{1-d}{2 N}\| \vec{\theta} \|^2 \\
& + \beta~\text{KL}[p_{\phi}(\vec{z_y}|\vec{X})|| p(\vec{z_y}|\vec{X},\vec{Y})],
\end{split}
\end{equation}
where $d$ is the dropout rate. 

\subsection{Computing Epsitemic and Labels Uncertainty}
To capture the \textit{labels} uncertainty (LU), $T$ stochastic forwarded passes are computed for a given sample $\vec{x}_i$ at a fixed weight estimate $\hat{\vec{w}_i}$ to calculate the predictive mean
\begin{equation}
\mathbb{E}_{LU}[\hat{\vec{y}_i}] = \frac{1}{T}\sum_{t=1}^{T} \vec{f}^{\hat{\vec{w}_i}}(\vec{x}_i, \vec{z_y}_i^t), 
\end{equation}
and the predictive variance
\begin{equation}
\text{Var}_{LU}[\hat{\vec{y}_i}] \approx \frac{1}{T}\sum_{t=1}^{T} \vec{f}^{\hat{\vec{w}_i}}(\vec{x}_i, \vec{z_y}_i^t)^T \vec{f}^{\hat{\vec{w}_i}}(\vec{x}_i, \vec{z_y}_i^t) - \mathbb{E}_{LU}[\hat{\vec{y}_i}]^T\mathbb{E}_{LU}[\hat{\vec{y}_i}].
\end{equation}
The \textit{epistemic} uncertainty (EU) is computed in a similar fashion, however, at a fixed latent representation $\vec{z_y}_i$ for a given input, and sample $\hat{\vec{w}_i}^t$ for $T$ times.
\section{Experiments and Results}
\label{sec:exp}

\paragraph{Datasets preparation.} We have mainly focused on classification task and evaluated the model first on MNIST as a proof-of-concept, and afterwards on a publicly available medical dataset, i.e. CheXpert\cite{irvin2019chexpert}. Since MNIST does not contain labels from multiple experts, the inter-observer variability was simulated. It was assumed that annotations of four experts are available, by replicating the ground truth for each of the assumed experts. For each replication, a certain part of the provided labels was changed to simulate different scenarios. The parameters such as the number of experts, the percentage of changes classes, and the number of changed classes are validated for MNIST dataset. 

\paragraph{Implementation.} Our proposed model consists of two main parts: a neural network responsible for the classification task $p(\vec{y}_i|\vec{f}^{\hat{\vec{w}_i}}(\vec{x}_i, \vec{z_y}_i))$ and a conditional variational autoencoder which models the inter-observer variability in its bottleneck $p_{\phi}(\vec{z_y}|\vec{X})$. Further details are provided in the Appendix. 
During training, $\vec{z_y}$ is sampled from the latent space of the posterior encoder $p(\vec{z_y}|\vec{X}, \vec{Y})$, tiled to match the dimension of the last feature vector of the classification network and concatenated in the channel axis, right before passing to the softmax layer. During testing, the procedure remains similar, with the only difference being that $\vec{z_y}$ is sampled now from the prior encoder $p_{\phi}(\vec{z_y}|\vec{X})$.

\paragraph{Proof-Of-Concept Experiments.} 
The proof-of-concept experiments are conducted on MNIST dataset, where $25\%$ of classes $2$ and $5$ are inter-exchanged for three out of four experts. The baseline model, trained on the untouched MNIST, achieves an accuracy of $99.2\%$ on the testing set. However, the accuracy drops to $97.1\%$ when the model trained on the simulated ones. 
The results of the uncertainties are reported in Table.\ref{mnist}. For labels uncertainty, we observed almost zero predictive variance on unchanged classes in both training and testing sets, while significantly higher uncertainty is reported for changed classes. The epistemic uncertainty exhibits similar behavior, only with our implementation setup, but with remarkable high variance, yielding an unreliable metric. 

\paragraph{Clinical Use-Case.} We evaluated our method on a recently released dataset for multi-label Chest X-ray classification, namely CheXpert \cite{irvin2019chexpert}, which contains some uncertain labels in its training set. It consists of 224,316 chest X-ray images of 65,240 patients. Each of them was labeled as positive, negative or uncertain for the occurrence of 14 observations by an automated rule-based labeler. Due to high class imbalance present in the dataset, we restrain our experiments to 6 most numerous classes out of 14. Classes of Consolidation and Atelectasis show a high percentage of uncertain samples, 12.78\% and 15.66\%, respectively. We chose uncertain samples from these two classes and doubled them treating the first part as positive and the second part as negative. The uncertain samples from the remaining four classes were completely removed from the dataset to constrain the problem. As a result, we obtain 106,993 data points in the training set. Dataset prepared in such a way contains a controlled amount of uncertainty that is possible to evaluate.

\textbf{Model Performance}
Similar to what we observed in MNIST experiments, both our model and the baseline perform worse in terms of AUC after adding uncertain data points. For instance, average AUC drops from 0.71 to 0.52 for the baseline and to 0.51 for our model. Interestingly, our proposed method reports a considerably higher predictive variance of classes containing uncertain samples than of the rest of the dataset. At the same time, it is not in any way reflected in the epistemic uncertainty that two classes contain data points labeled both as positive and negative. On the contrary, the two classes that would be expected to have the highest uncertainty yield the lowest score. Our results confirm our hypothesis that measuring epistemic uncertainty does not sufficiently capture the inter-observer variability.

\begin{table}[t]\label{mnist}	
\centering
\caption{Mean and (Variance) of uncertainty values for MNIST dataset.}
%\bigskip
\begin{tabular}{llllll}
 & Training set                                                                 &                 &  & Testing set       &                  \\ 
\hline
                                                                    & Unchanged classes & Changed classes &  & Unchanged classes & Changed classes  \\ 
\hline
%LeNet & & & & \\
LeNet$_{EU}$ & 0.0240 ($\pm 0.0127$)  & 0.5237 ($\pm 0.0968$) &&   0.1559 ($\pm 0.0124$) &   0.0739 ($\pm 0.0966$) \\
                                                                    
\hline
%Ours & & & & \\
\begin{tabular}[c]{@{}l@{}}Ours$_{EU}$\end{tabular}      & 0.0304 ($\pm 0.0650$)            & 0.5502 ($\pm 0.5789$)          &  & 0.1167 ($\pm 0.5567$)              & 0.5933 ($\pm 0.7476$)           \\
\begin{tabular}[c]{@{}l@{}}Ours$_{LU}$\end{tabular} & 0 ($\pm 0$)                                                                         & \textbf{0.0691} ($\pm 0.1087$)          &  & 0.0071 ($\pm 0.0465$)            & \textbf{0.0683} ($\pm 0.3614$)          \\
\hline
\end{tabular}
\end{table}
\begin{table}[t]
\centering
\caption{LU captures the uncertainty in both \textit{Consolidation} and \textit{Atelectasis} classes in CheXpert.}
%\bigskip
\begin{tabular}{lllllll}
\hline                                                                             \
Uncertainty  & Lung & Edema & Consolidation & Atelectasis & Pleural & Support\\
\hline
\begin{tabular}[c]{@{}l@{}}Ours$_{EU}$ \end{tabular}      & 0.0002       & 0.0009 & 0.0001        & 0.0001      & 0.0024           & 0.0011           \\
\begin{tabular}[c]{@{}l@{}}Ours$_{LU}$ \end{tabular} & 0.0046       & 0.0009 & \textbf{0.0283}        & \textbf{0.0609}      & 0.0012           & 0.0021           \\
\hline
\end{tabular}
\end{table}

\section{Conclusion}
\label{sec:conc}
In this paper, we introduce a novel framework for modeling labels and epistemic uncertainties capturing both the inter-observer variability and the model's weight uncertainty, respectively. Our proof-of-concept experiments, together with the clinical use-case on a publicly available dataset, confirm our hypothesis of the essential need of modeling both types of uncertainties. Our results show a high potential of utilizing our framework for quality control and referral systems. 

\section*{Acknowledgment}
The work leading to this publication was supported by the PRIME programme of the
German Academic Exchange Service (DAAD) with funds from the German Federal Ministry
of Education and Research (BMBF).

\bibliography{LabelsUncertainty,biblio-macros}
\bibliographystyle{splncs03}

%\newpage
%\section*{Appendix}
%\input{appendix/impl}
%\input{appendix/results}
\end{document}